\documentclass[runningheads]{llncs}
\usepackage{graphicx}
\usepackage{amsmath,amssymb} % define this before the line numbering.

\begin{document}
%===========================================================

\title{Neural Abstract Style Transfer for Chinese Traditional Painting} % Replace your paper's title here
\titlerunning{Neural Abstract Style Transfer for Chinese Traditional Painting} % Replace an abstracted version of your paper's title here

%===========================================================

\author{Bo Li\inst{1} \and
Caiming Xiong\inst{2} \and
Tianfu Wu\inst{3} \and
Yu Zhou\inst{4} \and
Lun Zhang\inst{1} \and
Rufeng Chu\inst{5}}
%
%Please include author names in full in the paper, 
%If any authors have names that can be parsed into FirstName LastName in multiple ways, please include the correct parsing, in a comment to the volume editors:
%\index{Lastnames, Firstnames}

\authorrunning{B. Li et al.} % A shorter version of authors' name

%===========================================================

\institute{Alibaba Group, Beijing, China \and
Salesforce Research, CA, USA \and
Department of ECE, NC State University, NC, USA \and
Beijing University of Posts and Telecommunications, Beijing, China \and
Beijing Winsense Technology Co. Ltd, Beijing, China \\
\email{\{shize.lb,lunzhangl\}@alibaba-inc.com, cxiong@salesforce.com, tianfu\_wu@ncsu.edu, yuzhou@bupt.edu.cn, rfchu@winsense.ai}}

\maketitle

%===========================================================
\begin{abstract}
Chinese traditional painting is one of the most historical artworks in the world. It is very popular in Eastern and Southeast Asia due to being aesthetically appealing. Compared with western artistic painting, it is usually more visually abstract and textureless. Recently, neural network based style transfer methods have shown promising and appealing results which are mainly focused on western painting. It remains a challenging problem to preserve abstraction in neural style transfer. 
%For example, the model of \cite{gatys_cvpr16} usually computes a ``shallow" color-filtered image (as many Chinese traditional painting are less colorful) with the ``abstract" style of the target Chinese traditional painting  not captured well. 
In this paper, we present a Neural Abstract Style Transfer method for Chinese traditional painting. It learns to preserve abstraction and other style jointly end-to-end via a novel MXDoG-guided filter (Modified version of the eXtended Difference-of-Gaussians) and three fully differentiable loss terms. 
To the best of our knowledge, there is little work study on neural style transfer of Chinese traditional painting.
To promote research on this direction, we collect a new dataset with diverse photo-realistic images and Chinese traditional paintings\footnote{The dataset will be released at https://github.com/lbsswu/Chinese$\_$style$\_$transfer.}.
In experiments, the proposed method  shows  more appealing stylized results in transferring the style of Chinese traditional painting than state-of-the-art neural style transfer methods.

\end{abstract}
%===========================================================
\section{Introduction}

Chinese traditional painting is an ancient art form, in which natural objects are painted with sparse, yet expressive, brush strokes. It consists of diverse styles (e.g, claborate-style painting, Chinese landscape painting, and ink and wash) and has influenced many countries and nations in Eastern and Southeast Asia. 
It's now a typical symbol of Chinese culture and an important part of the artistic world. 

Recently, convolutional neural network (CNN) \cite{lecun98} based style transfer methods have shown successful applications in transferring the style of a certain type of artistic painting, e.g, Vincent van Gogh's ``The Starry Night", to a real world photograph, e.g., an image taken by iPhone. Since the seminal work of Gatys et al. \cite{gatys_cvpr16}, it has attracted a lot of attentions from both academia \cite{johnson_eccv16,lichuan1,nikulin1,nikulin2,gatys_cvpr17,portrait,ulyanov,chen2018stereoscopic,chen2017stylebank,chen2017coherent} and industry \cite{prisma,artisto,kristen,museum}. 
Although the work of neural style transfer has shown promising progress on transferring artistic images with rich textures and colors, e.g., the oil paintings, we observe that it is less effective in transferring Chinese traditional painting.
 
Unlike western oil paintings which are often concrete and realistic, Chinese traditional freehand painting reveals an artistic results of a likeness in spirit rather than in appearance. As a result, different styles of sparse brush strokes are widely utilized to depict different kinds of objects. Thus they are more abstract, textureless and less colorful. And this ``abstract style" is not captured well by current neural style transfer methods due to  lack of corresponding constraints.

%\noindent 
\begin{figure*}
\centering
%\framebox
{\includegraphics[width = 1.0\textwidth]{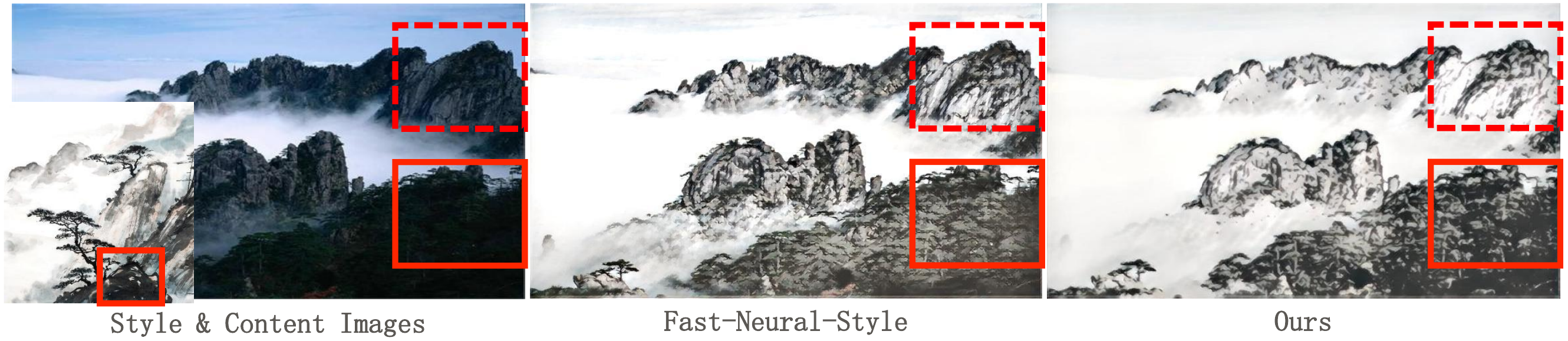}}
\caption{Stylized examples of neural style transfer \cite{gatys_cvpr16} and our method for Chinese traditional painting. The left column shows the input content image and style image. The middle column shows the transferred result of the neural style transfer method \cite{gatys_cvpr16}. The right column shows the stylized result of our proposed method. From which we can see, the result generated by our method is more sparse and the style is more like the style image. }
\label{fig:demo}
\end{figure*} 
%Fig. 1 Stylized example of neural style methods on Chinese traditional artistic painting.

Fig. \ref{fig:demo} shows an example. The left figure shows an input real image superposed with a Chinese traditional painting as target style. The middle figure shows the stylized result of neural style transfer \cite{johnson_eccv16,gatys_cvpr16} which does not capture the abstract style as concise and clean as the target style image. For instance, trees (solid rectangle) and mountains (dashed rectangle) are not transferred very well, as there are still many redundant edges or stokes on them, which should be abstracted out w.r.t. the style image.
Besides, strokes in the stylized results do not align with those in the style image. For example, the style of strokes in the dark area (solid rectangle in the middle figure) stylized by Fast-Neural-Style \cite{johnson_eccv16}  is still quite different as the one (solid rectangle in the left figure) in the style image, making these areas looks trivial and non-smooth.
These comparisons make it clear that we need to learn to ``abstract" and keep a smooth and natural transfer that consistent with the style of Chinese traditional painting. This issue has not been addressed in existing methods.

In this paper, we focus on the specific and important problem of style transfer of Chinese traditional painting.
Then, to address above issues of current neural style transfer methods, we propose a modified extended difference-of-Gaussians (MXDoG)  based style transfer approach for Chinese traditional painting, where a MXDoG filter is utilized to abstract an image. Based on the MXDoG,  we formulate three new terms in the loss function for neural style transfer beside the conventional content loss and style loss. 

The first loss term is a MXDoG content loss, which penalizes the discrepancy of appearance between the stylized image and the MXDoG filtered image.  We suppose the representation of the \textbf{abstract content} of an image is also separable along with the \textbf{content} and the \textbf{style} of the same image, and this loss term will impose a new constraint that requires the stylized image to have a  ``balanced content" that accommodating to both the ``content" and the ``MXDoG abstracted content" of  an image.
The second loss term penalizes the dissimilarity between the MXDoG filtered image of the stylized image and the content image. It is inspired by the work of \cite{lapstyle} which uses the Laplacian operator.
The third loss term focuses on style, which encourages that the MXDoG filtered image of the stylized image and the style image to have similar styles.
The second and third terms are mainly used to penalize large noisy edges in stylized images to make the result more natural. 
%As we care more about the abstract part rather than the very low-level details, our MXDoG is more preferable than the Laplacian filter \cite{lapstyle} in style transfer of Chinese traditional painting.
These three loss terms are fully differentiable, thus our style transfer network can be trained end-to-end by stochastic gradient descent method. 

An example of our stylized image is shown in Fig. \ref{fig:demo}(c). Overall, our model shows more appealing style which respects the target style image than neural style transfer methods. 
For example, our model produces less strokes for the mountain peak in the dashed rectangle than Fast-Neural-Style \cite{johnson_eccv16}. This is more in accord with Chinese traditional painting in terms of sparse strokes.
In addition, for the dark area (solid rectangle) in the content image, our stylized result is more in accord with the dark area in the style image.

It's worth noted that our method is not necessarily only applicable to Chinese traditional painting. The proposed three new loss terms are used for handling the abstractness and textureless in style transfer, since artworks (e.g., ukiyoe, cartoon, oil painting) have different extents of abstraction, it can work for general art styles by adapting the hyper-parameters of these loss terms. Automatically learning the hyper-parameters is very attractive, we leave it as an interesting future work and focus our efforts on transferring Chinese traditional painting.

%Extended difference-of-Gaussians [] is less susceptible to noise and is aesthetic appealing, it is able to to yield more aesthetically pleasing edge lines than edges\footnote{Edges extracted by an edge detector, e.g., the Canny edge detector [], require significant postprocessing to adapt to artistic stylization.} (see Fig. 2). From Fig. 2, we can see XDoG version of an image provide some ``abstraction", this character is very useful in synthesizing line drawings and cartoons [], and it is also useful in traditional Chinese art works, as there are also many freehand edge lines. 

To the best of our knowledge, there is no publicly available dataset for evaluating Chinese traditional painting style transfer,  thus we collect a new dataset that contains a variety of natural scenes and Chinese traditional paintings. The dataset will be released to facilitate further research on this direction.

In experiments, we compare our method with the neural style methods \cite{gatys_cvpr16,lichuan1,johnson_eccv16} on transferring the style of Chinese tradition paintings, and show that our method performs better on transferring image textures, abstract contents, and colors. In addition,   the stylized images are ``clean", natural and have strong layers of graphics.

We make the following contributions to the community of image style transfer:
\begin{itemize}
\item We reintroduce the problem of style transfer of Chinese traditional painting, which poses new challenges and largely omitted by current research. %To the best of our knowledge, this is the first end-to-end neural style transfer work focus on  Chinese traditional painting.
\item We propose a  MXDoG filter to abstract the content of an image, and utilize it to transfer the style of Chinese traditional painting.
\item We propose three MXDoG based loss terms to guide the neural networks to learn how to ``abstract", and demonstrate its effects on test images under different conditions. In this way, we also verify the representations of ``abstract content", ``content" and ``style" of an image can be separated by the neural networks.
\item We collect a new  Chinese traditional painting dataset to promote the research on style transfer of Chinese traditional painting. 
%The dataset will be released  to public to facilitate further research.
\end{itemize}

\section{Related Works}
We briefly review related works of neural style transfer and style transfer of Chinese traditional painting below.

\textbf{Neural Style Transfer.}  Gatys et al. \cite{gatys_cvpr16} first propose the neural network based style transfer method, in which they synthesizing images that have the style of one image and the content of another. In their method, the style is represented by the Gram matrix,  and the content is represented by high-level convolutional feature maps.
Here, the Gram matrix is the global statistics of the image based on outputs from convolutional layers.
Gatys's work has received lot of attentions and triggered a whole line of research on deep learning based style transfer. \cite{nikulin1,nikulin2} investigate several variants of Gatys' method for illumination and season transfer.
Li et al. \cite{lichuan1} utilize the patch-based Markov random field method to represent the style of the image with neural networks. Luan et al. \cite{luan2017deep} propose a method for photo to photo style transfer, which shows high quality in photo-realistic.

Recently, Li et al. \cite{lapstyle} introduce a Laplacian loss term to preserve detailed content image structures.
The key difference between our work and \cite{lapstyle} is XDoG vs LoG, rather than DoG vs LoG. DoG (Difference-of-Gaussians) is a fast approximation of the LoG (Laplacian of Gaussians), while XDoG is built on DoG/LoG which detects edges by thresholding DoG responses, rather than searching for the zero crossings in the second derivative (see Eqn. (6)).
XDoG is more aesthetically appealing than DoG/LoG due to its effects on \textbf{edge enhancement}. Edge enhancement focuses more appropriately on the weight (thickness)
and structure (shape) of edges, thus providing better results for stylistic and
artistic applications \cite{xdog}.

\textbf{Fast Neural Style Transfer.} Above neural style transfer methods utilize optimization for image style transfer, usually, it takes more than $40$ seconds to process an image. Johnson et al. \cite{johnson_eccv16} utilize the perceptual loss to train feed-forward neural networks, which can be running in real time on GPU. Almost at the same time, Ulyanov et al. \cite{ulyanov} propose an unsupervised real time method, but  a multi-scale neural network is used. Li and Wand \cite{lichuan2} also propose a feed-forward method to accelerate their patch-based Markov method \cite{lichuan1}. 
Recently, Ulyanov et al. \cite{inst_norm} further propose an instance normalization method which significantly improves the quality of fast neural style transfer.

\textbf{Style Transfer with GAN.} Recently, several work \cite{cycle_gan,lichuan2,auto_painter,filamentary} try to use or incorporate generative adversarial networks (GAN) for image style transfer. Specifically, the Cycle-GAN method \cite{cycle_gan} produces amazing results in transferring an image with a painter's style, e.g., Vincent van Gogh, Monet. However, this method is not stable, needs much more time and requires large number of unpaired content and style images for training.
What's more, the style of a painting maybe quite different from another even they are painted by the same artist, thus it may be not desirable when we just want to transfer the style of a specific artwork.

\textbf{Style Transfer for Chinese traditional painting.} Before deep neural network is prevalent, many researchers \cite{strassmann,zhangsonghai,way2002,xu2006animating,lee1999simulating,baxter2001} focus on simulation of the interaction of water, ink, paper and brushes to render Chinese tradition painting. Recently, \cite{Chinese_painting_lindaoyu} propose to transfer Chinese painting using multi-scale neural network, however, their method is not end-to-end, and requires sketches or edges for input. Overall, there is little work specifically for style transfer of Chinese traditional painting, thus  challenges of style transfer of Chinese traditional painting are largely omitted by our community. In this paper, we make an preliminary analysis of these challenges, and hope more researchers will join and promote the research on this direction.

\section{Method}
We first briefly review the neural style transfer method, then we introduce the modified extended  difference-of-Gaussians (MXDoG) filter, which produces a novel representation for image abstraction in our framework.
Based on MXDoG, we further introduce the  structure and loss functions of our neural network architecture.

\subsection{Neural Style Transfer}
Given a content image $I_c$ and a style image $I_s$, the goal of image style transfer is to generate an image $I$ showing the content of $I_c$ in the style of $I_s$. Gatys et al. \cite{gatys_cvpr16} formulate the image style transfer as an energy minimization problem which consisting of a content loss and a style loss. Both losses are computed with an ImageNet pretrained object classification network (i.e. VGG-19 \cite{vgg}). 

Inputing an image $I_c$ to the pre-trained network, we can  get the $l-$th  feature map $F_l(I)=\phi^l(I)$ which corresponds to the response of the $l-$th layer. The dimension of $F_l(I)$ is $N_l \times M_l(I)$, where $N_l$ is the number of filters (channels) in the $l-$th layer, and $M_l(I)=H_l(I) \times W_l(I)$ is the spatial dimension of the $l-$th feature map, i.e. the product of its height and width.

With above notations, the objective of neural style transfer method can be represented as follows:
\begin{eqnarray} \label{eqn:neural_obj}
& L_{T}(I, I_c, I_s) = \alpha * L_{C}(I, I_c) + \beta * L_{S}(I, I_s)
\end{eqnarray}
where $\alpha$ and $\beta$ are the weighting factors showing the relative importance of the two components, the content loss is the mean-squared distance between the feature map of $I_c$ and $I$ at a specified layer $l$:
\begin{eqnarray} \label{eqn:content}
& L_{C}(I, I_c) = \frac{1}{N_l M_l(I_c)} \sum_{ij}(F_l(I) - F_l(I_c))_{ij})^2
\end{eqnarray}
and the style loss is the mean-squared distance between the correlations of the filter responses (i.e., Gram matrices) of $I_s$ and $I$ at several appointed layers:
\begin{eqnarray} \label{eqn:style}
& L_{S}(I, I_s) = \sum_l \frac{\sum_{ij}(G^l_{ij}(I)-G^l_{ij}(I_s))^2}{N_l^2} 
\end{eqnarray}
where $G^l_{ij}(I)=\frac{1}{M_l(I)} \sum_{k=1}^{M_l(I)} \phi^l_{ik}(I)\phi^l_{jk}(I)$ is the Gram matrix of $F_l(I)$.
The stylized image is generated by iteratively minimizing Eqn. (\ref{eqn:neural_obj}).

Instead of solving an optimization problem, Johnson et al. \cite{johnson_eccv16} propose a much faster feed-forward network to directly mapping an input image to the  stylized one, this method is called Fast-Neural-Style transfer. Denote the parameters of the feed-forward network as $w$, the training objective is as follows:
\begin{eqnarray}
& w^* = arg min_w E_I[L_{T}(I, I_c, I_s)]
\end{eqnarray}
where  $E$ is the expectation.

\subsection{Modified Extended  Difference-of-Gaussians}

\begin{figure}
\centering
%\framebox
{\includegraphics[width = 0.96\textwidth]{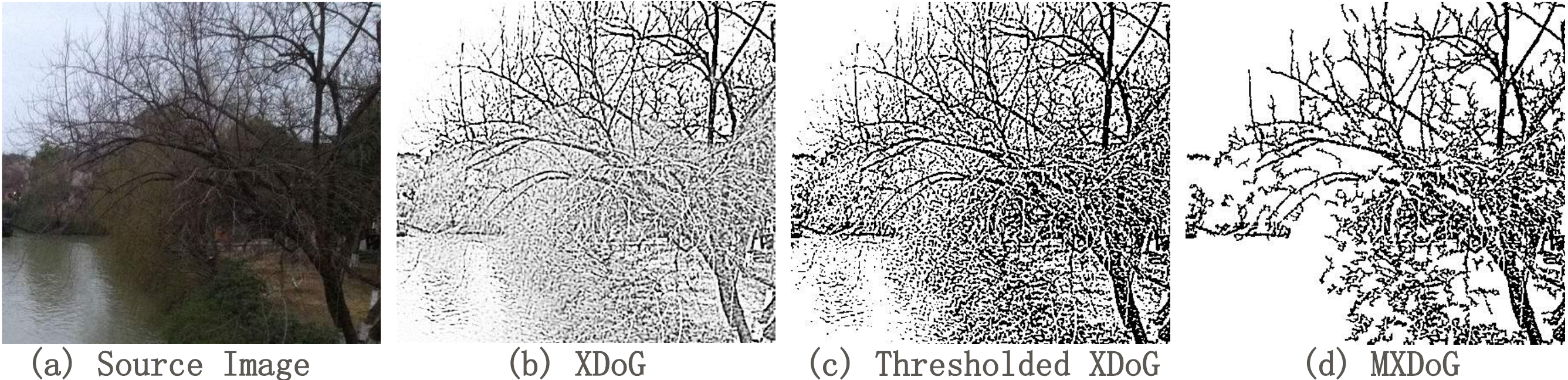}}
\caption{Filtered results of XDoG, Thresholded XDoG, and MXDoG.}
\label{fig:xdog}
\end{figure}

The extended difference-of-Gaussians (XDoG) operators have been shown to yield a range of subtle artistic effects, such as ghosting, speed-lines, negative edges, indication, and abstraction etc \cite{xdog}. 
Chinese traditional painting share some similar characters with above artistic paintings, e.g, abstraction, textureless, emphasis of edges.
Thus XDoG filters are  attractive for us in improving the quality of style transfer for Chinese traditional painting.

%Note XDoG is not an edge detector in a strict view. To see this we show the result of Canny edge detector in Fig. \ref{fig:xdog}(b). Comparing Fig. \ref{fig:xdog}(b) and Fig. \ref{fig:xdog}(c), we can see the XDoG is more attractive from an artistic point of view.
%From an artistic point of view, the results of the Canny detector are rather unattractive, as illustrated in Fig. \ref{fig:xdog}, and adopting them to the task of artistic stylization requires significant post-processing. 

Given an image $I$,  traditional XDoG filter can be formulated as:
\begin{eqnarray}
& I^{xd} = \mathcal{T}_{\varepsilon, \varphi}(\mathcal{D}_{\sigma,k,\tau}(I))
\end{eqnarray}
where $\mathcal{T}$ is the XDoG filter and $\mathcal{D}$ is a variant of the difference-of-Gaussians filter in \cite{xdog}.
$\mathcal{T}$  can be formulated by a thresholding funtion with a continuous ramp:
\begin{eqnarray} \label{eqn:xdog}
& \mathcal{T}_{\varepsilon, \varphi}(u) = 
\begin{cases}
1 & \text{$u \geq \varepsilon$} \\
1 + tanh(\varphi \cdot (u - \varepsilon)) & \text{otherwise}.
\end{cases}
\end{eqnarray}
where $\varphi$ and $\varepsilon$ are the related thresholding parameters. And $\mathcal{D}$ can be formulated as 
\begin{eqnarray} \label{eqn:dog}
& D_{\sigma,k,\tau}(x) = g_{\sigma}(x) - \tau \cdot  g_{k \sigma}(x) 
\end{eqnarray}
where $ g_{\sigma}(x) = \frac{1}{2 \pi \sigma^2} exp(-\frac{\parallel x \parallel^2}{2 \sigma^2})$ is the Gaussian smoothing filter, $k$ represents a trade-off parameter balancing accurate approximation and adequate sensitivity \cite{edge_theory}, $\sigma$ is the standard deviation and $\tau$ is the control parameter.

Traditional XDoG is aesthetically appealing and can abstract an image to some extent (see Fig \ref{fig:xdog}(b)). However, it's still not enough for general natural images, as there are many small pieces in the image (which is still too detailed). In addition, the XDoG processed image is generally too ``white" and is not very compatible with the style of Chinese traditional painting (as the contrast of black and white colors in  Chinese traditional painting is generally striking).
To this end, we propose a novel modified XDoG (MXDoG) that is a thresholded version of XDoG, and incorporate morphology operations to filter out the small pieces in an image. 
Our MXDoG is formulated as:
\begin{eqnarray} \label{eqn:mdog}
& I^{md} = morph\_filter(I^{td})
\end{eqnarray}
where $morph\_filter$ is the morphology operation which filtering out image regions with their areas smaller than a predefined minimum size $A_{min}$, and $I^{td}$ (Fig. \ref{fig:xdog}(c)) is the thresholded XDoG which is formulated as:
\begin{eqnarray} \label{eqn:thresh_xdog}
& I^{td}(x) = 
\begin{cases}
0 & \text{$I^{xd}(x) \leq \mu$} \\
1, & \text{otherwise}.
\end{cases}
\end{eqnarray}
where $\mu$ is the mean of $I^{xd}$.

For a color image, we compute MXDoG for each channel separately.
The final result of our MXDoG operator on a sampled image is shown in Fig. \ref{fig:xdog}(d).

%It should be pointed out that the  abstraction mechanism used by the XDoG filter has significant limitations and does not compare with the skill of a trained artist. 
%For example, there are still some unimportant edges and  spotted patterns in the XDoG processed image. However, as {\it abstraction} remains some of the fundamentally unsolved problems in non-photorealistic rendering (NPR) [], this arguably good results still  help a lot for our neural stylized model to get a freehand painting and might steer deeper research into the artistic neural style transfer problem.

\subsection{Network Architecture}

\begin{figure*}
\centering
%\framebox
{\includegraphics[width = 0.90\textwidth]{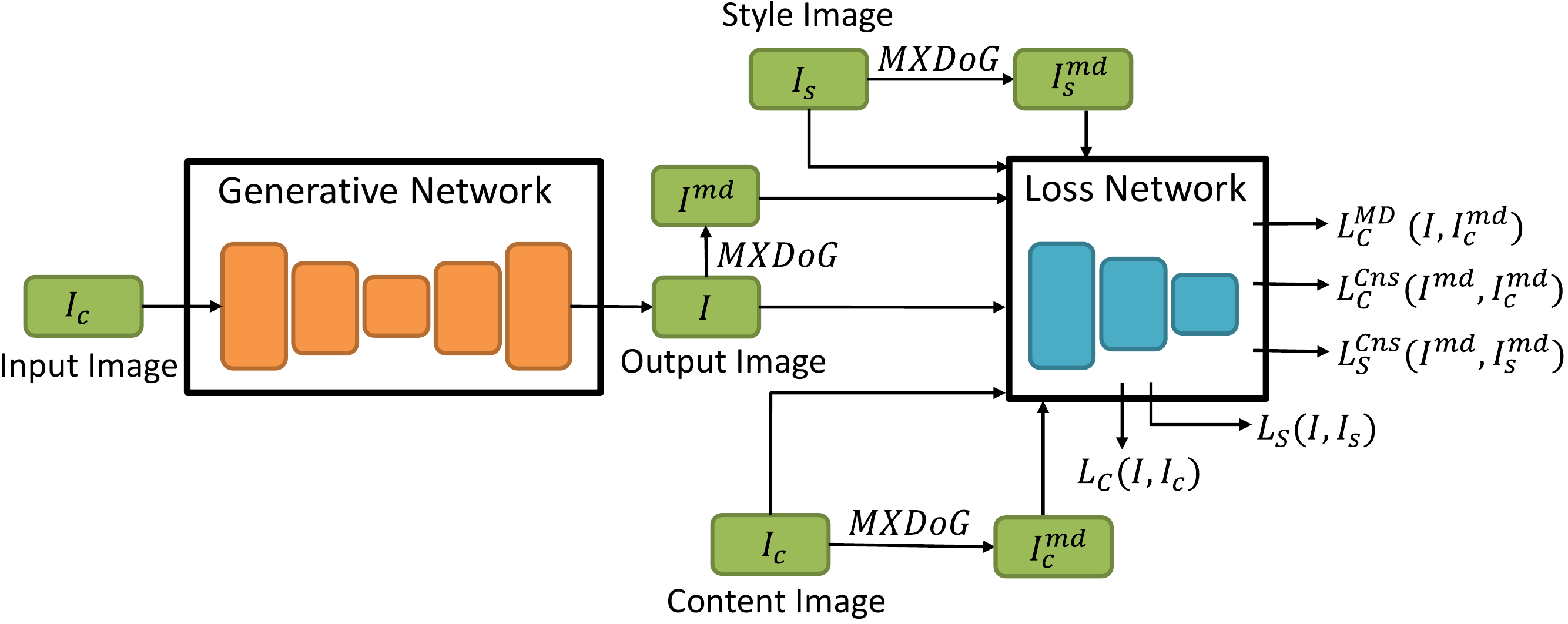}}
\caption{Network architecture including a generative network and a loss network. As a common practice, we use the content image $I_c$ as the input image to do style transfer.}
\label{fig:model}
\end{figure*} 

Our style transfer system consists of two components: a {\it generative network}  and a {\it loss network}, as illustrated in Fig. \ref{fig:model}. The generative network is responsible for transforming a user-provided image $I_c$ (\textbf{as a common practice, we use the content image $I_c$ as the input image}) to a corresponding stylized image. It is a deep residual convolutional neural network with similar network structure as \cite{style_wenhan}. 
The loss network is an ImageNet \cite{imagenet} pretrained object classification network, and is fixed during the training of the generative network.  Throughout this paper, we use the $16$-layer VGG network \cite{vgg} as the loss network. 
 Besides the content loss and style loss used in  \cite{gatys_cvpr16} and  \cite{johnson_eccv16}, we  introduce three new MXDoG-based losses. The generative network is trained using stochastic gradient descent to minimize the following overall loss function:
\begin{align} \label{eqn:obj}
& \nonumber  L(I,I_c,I_s,I^{md},I_c^{md},I_s^{md}) = \lambda_1 L_{C}(I,I_c) + \lambda_2 L_{C}^{MD}(I,I_c^{md}) + \\
&  \lambda_3 L_{S}(I,I_s) + \lambda_4 L_{C}^{Cns}(I^{md},I_c^{md}) + \lambda_5 L_{S}^{Cns}(I^{md},I_s^{md})
\end{align}

Here $\lambda_i$ is the weighting factors to combine various loss components. 
The content loss $L_C$ is computed as Eqn. (\ref{eqn:content}).
It is worth noted that although the content loss provides some extent of ``abstraction", it is still not enough (see Fig. \ref{fig:demo}).
The style loss $L_{S}$ is computed as Eqn. (\ref{eqn:style}). As the styles of traditional Chinese artworks are often textueless and lack of color information, we strengthen the effects of style transfer by using more low-level and high-level layers for style reconstruction than  \cite{johnson_eccv16}.
The implementation details will be exposed in our experiment.
Details of our proposed three new loss terms (i.e., $L_C^{MD}$, $L_C^{Cns}$, and $L_S^{Cns}$) are as follows:

\subsubsection{MXDoG Content Loss.}
Given content image $I_c$, we utilize the MXDoG filter to produce the abstract content image $I_c^{md}$.
Then  $I_c^{md}$ is used as the input  to the loss network to extract high-level features of VGG-16 net. Similar to the content loss, we compute mean-squared Euclidean distance between feature representations $F_l(I)$ and $F_l(I_c^{md})$ as:
$L_{C}^{MD}(I, I_c^{md}) = L_{C}(I, I_c^{md})$,
%\begin{eqnarray}
%& \nonumber L_{C}^{MD}(I, I_c^{md}) = \frac{\sum_{ij}(F_l(I) - F_l(I_c^{md}))_{ij})^2}{N_l M_l(I_c^{md})} 
%\end{eqnarray}
In the experiments, we use the same mid-level layer, i.e., {\it relu3\_3}, for both the content loss and XDoG content loss.

$L_{C}^{MD}$ penalizes the output image $I$ when it deviates in content from the target $I_c^{md}$. In other words, $L_C^{MD}$ asks the output image $I$ to have similar appearances with the MXDoG filtered content image $I_c^{md}$. Here $I_c^{md}$ is used as the ``abstract content image". 

By providing two content loss: a concrete one $L_{C}(I, I_c)$ and an abstract one $L_{C}^{MD}(I,I_c^{md})$, the generative network is encouraged to find a mid-point to balance the fidelity of the photorealistic appearance and the aesthetics of the artistic abstraction. 
The right figure in Fig. \ref{fig:demo} shows a sample result, we can see the generative network indeed learns how to discard some unimportant details (e.g., the mountain peak in the dashed rectangle) when compare with the one produced by the neural style transfer method (the middle figure in Fig. \ref{fig:demo}) which using the content loss only.

\subsubsection{MXDoG Content Constraint Loss.}
In the stylized image, there are often some noisy edges or distorted artifacts which is inconsistent with the content image. Inspired by \cite{lapstyle}, we introduce a new  loss that constrain $I^{md}$ to have similar appearances to $I_c^{md}$. This loss is defined as the mean-squared distance between $I^{md}$ and $I_c^{md}$, which drives the stylized image to have similar detail structures as the content image. This loss is dubbed as MXDoG content constraint loss as:
$L_{C}^{Cns}(I^{md},I_c^{md}) = L_{C}(I^{md},I_c^{md})$,
%\begin{eqnarray}
%& \nonumber L_{C}^{Cns}(I^{md},I_c^{md}) = \frac{\sum_{ij}(F_l(I^{md}) - F_l(I_c^{md}))_{ij})^2}{N_l M_l(I_c^{md})} 
%\end{eqnarray}
where $I_c^{md}$  and $I^{md}$ are computed by Eqn. (\ref{eqn:mdog}).
We use the layer  {\it relu3\_3} of VGG16 \cite{vgg} to get the mid-level patterns and impose the MXDoG content constraint on it.

 As MXDoG extracts the ``abstract content" of an image, this loss only penalizes the deviation of relatively larger edges or patterns instead of very detail fine structures.  This is different from the Laplacian loss used in \cite{lapstyle}.

\subsubsection{MXDoG Style Constraint Loss.}
In addition to the MXDoG content constraint, we also add a new  loss that constrain $I^{md}$ to have similar styles as $I_s^{md}$.
The motivation is if the styles of two images are similar, then the styles of their MXDoG filtered images are also similar.
%Our intuition is, since the output image $I$  have a similar style to the style image $I_s$, then the MXDoG filtered images $I^{MD}$ and $I_s^{MD}$ should also have similar styles. 
As similar to the style loss $L_S$, we compute the mean-squared error between the Gram matrices of $I^{md}$ and $I_s^{md}$:
$L_{S}^{Cns}(I^{md},I_s^{md}) = L_{S}(I^{md},I_s^{md})$,
%\begin{eqnarray}
%& \nonumber L_{S}^{Cns}(I^{md},I_s^{md}) =  \sum_l \frac{\sum_{ij}(G^l_{ij}(I^{md})-G^l_{ji}(I_s^{md}))^2}{N_l^2} 
%\end{eqnarray}
where $I_s^{md}$ is also computed by Eqn. (\ref{eqn:mdog}).
This loss further constrains the style consistence of the stylized image and the style image.

\begin{figure*}
\centering
%\framebox
{\includegraphics[width = 0.99\textwidth]{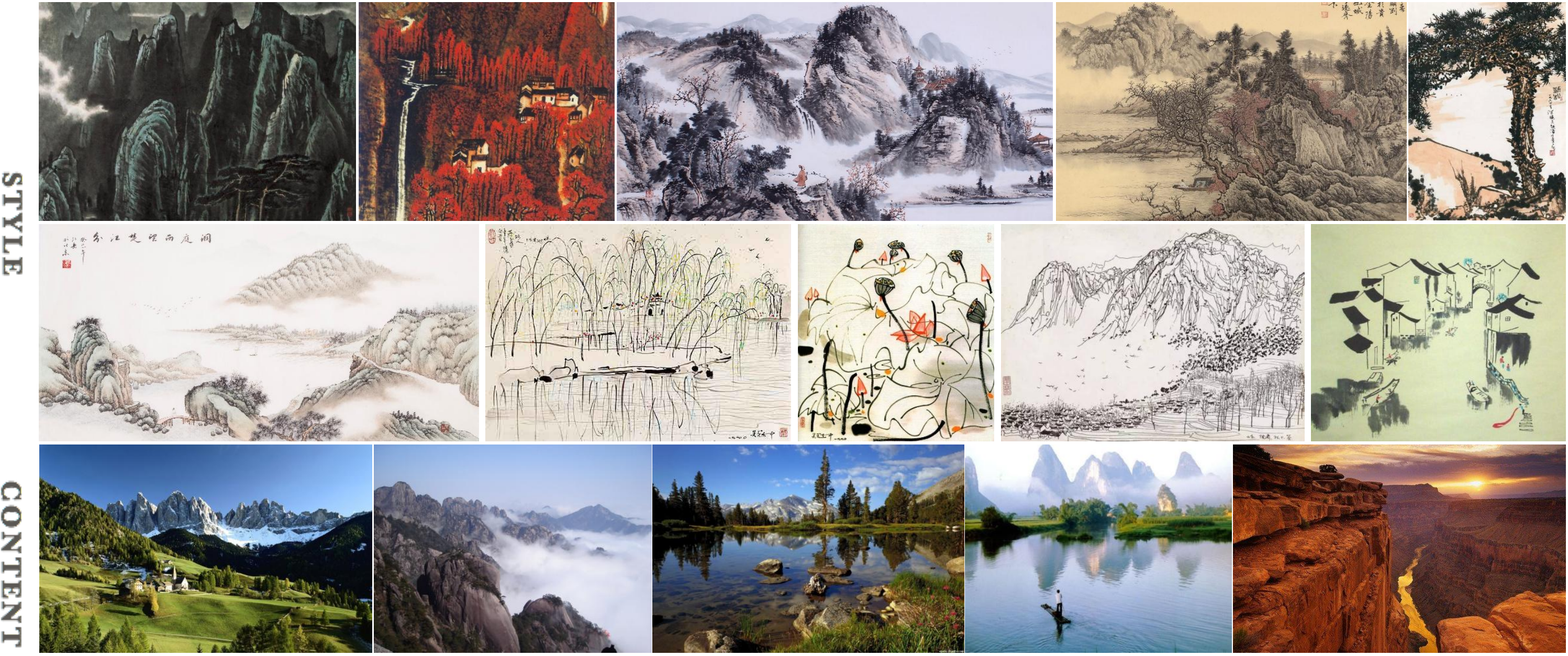}}
\caption{Sampled style and content images from the Chinese traditional painting dataset.}
\label{fig:dataset}
\end{figure*} 

\section{Results} 
%In this section, we first describe the implementation details of proposed method,  then we briefly introduce our self-collected Chinese traditional painting dataset. Next, we present the qualitative and quantitative comparisons of our method and state-of-the-art methods. At last, we present some typical failure examples of our method.

\subsection{Implementation Details} 
Our model builds upon the  fast neural style transfer framework of  \cite{johnson_eccv16}, which using the perceptual loss to train feed-forward neural networks to make the stylization achieving real time performance. we only introduces some computational burden on computing the MXDoG loss terms during offline training, and it doesn't adding any extra cost on online testing. 
The model is trained on the Microsoft COCO database \cite{coco}, which has around $80k$ training images. We resize these images to $256 \times 256$ and train our model using a batch size of $4$ for $2$ epochs.  We adopt Adam \cite{adam} for training with a learning rate of $1 \times 10^{-3}$.
Eqn. (\ref{eqn:obj}) is utilized as the loss function with the balancing weights $\lambda_1 = 1.0$, $\lambda_2 = 0.1 \sim 0.3$, $\lambda_3 = 5.0$, $\lambda_4 = 2 \times 10^2$ and $\lambda_5 = 1 \times 10^3$. For the computation of our MXDoG, we set $\tau = 0.94$, $\sigma=1.0$, $k=1.6$, $\varphi=50$, $\varepsilon = -0.1$ and $A_{min}=10$.
We compute the content loss at layer $relu3\_3$ and style reconstruction loss at layers $relu1\_2$, $relu2\_1$, $relu2\_2$, $relu3\_1$, $relu3\_3$, $relu4\_1$ and $relu4\_3$ of the VGG-16  loss network. 
All the parameters are chosen based on the MS-COCO 2014 validation set. We implement our method using PyTorch \cite{pytorch} with CUDA 7.5 and cuDNN 5.0. It takes about $8$ hours  to train a model with a single NVIDIA Tesla K$40$ GPU.
After training, our generative network can accept arbitrary input image size, and we resize the input image with the longer edge as $768$ before style transfer.

\subsection{Chinese traditional painting Dataset}
To the best of our knowledge, there is little work study on neural network based  Chinese traditional painting style transfer. Thus we collect one with  $1000$ content images that accommodate to the extent of   Chinese traditional painting. These images are collected by web search engines, e.g., Google, Baidu and Bing. They are mostly the photorealistic scenes of mountain, lake, river, bridge,  and buildings in regions south of the Yangtze River. It includes not only the scenes of China, but also beautiful pictures of Rhine, Alps, Yellow Stone, Grand Canyon, etc. These images are only used for testing.
Besides, we also collect $100$ traditional Chinese artworks. These artworks are used as the style images in this paper, which are the typical freehand brush works of China. 
Some typical style and content images of this dataset are presented in Fig. \ref{fig:dataset}.
The whole dataset including all the content and style images will be released to public for further research.

\subsection{Baselines}
To verify the capability of our model, we compare our method with state-of-the-art methods, i.e., Neural-Style Transfer by Gatys et al. \cite{gatys_cvpr16}, Fast-Neural-Style Transfer by Johnson et al. \cite{johnson_eccv16}, and CNN-MRF by Li and Wand \cite{lichuan1}.
As they all released their packages, we train their models by using the  Chinese traditional painting as style images for comparison. Details will be described in the following.

\subsection{Qualitative Results}

\begin{figure*}
\centering
%\framebox
{\includegraphics[width = 0.99\textwidth]{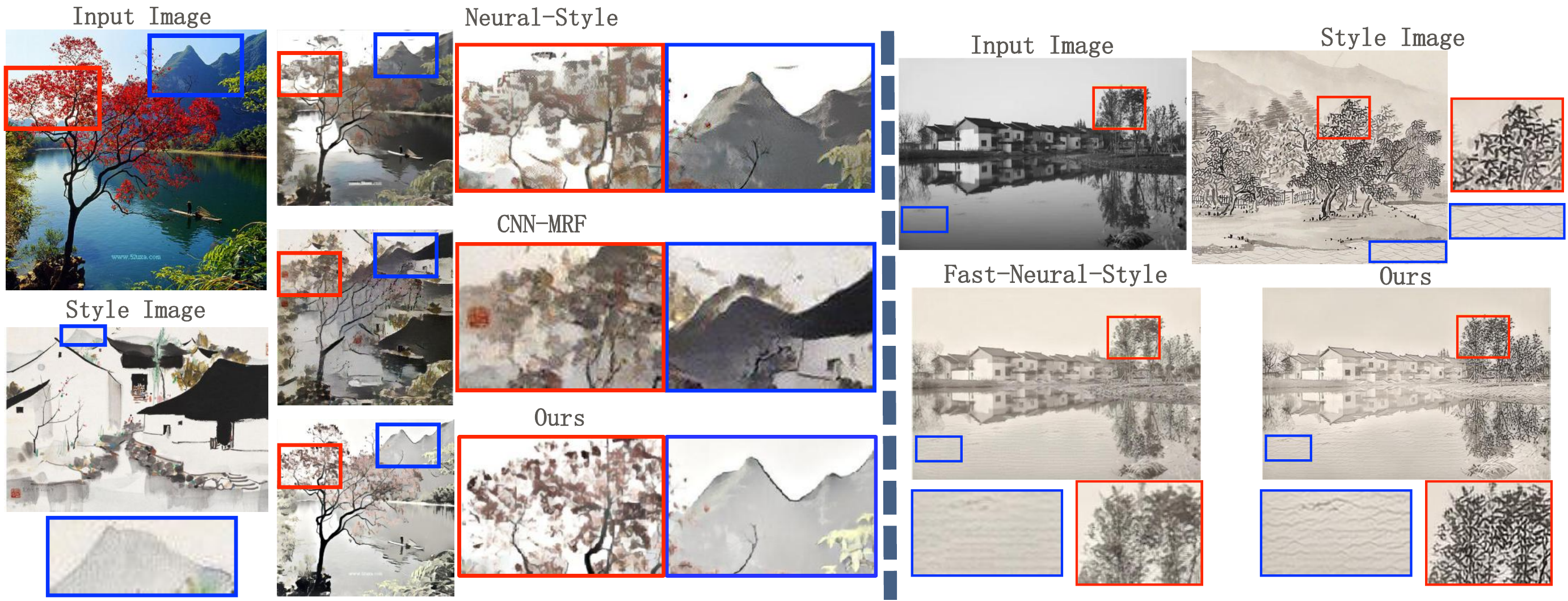}}
\caption{Left: Comparisons of our method and state-of-the-art Neural-Style \cite{gatys_cvpr16} and CNN-MRF  \cite{lichuan1}. Right: Comparisons of our method and Fast-Neural-Style \cite{johnson_eccv16}.}
\label{fig:qua}
\end{figure*}

We compare our method with state-of-the-art methods for a variety  of style and content images on  Chinese traditional painting dataset. On the left column of Fig. \ref{fig:qua}, we stylising a photograph with the same artwork for Neural-Style, CNN-MRF and our method. Result of each method is displayed on the right. To analyse the effect of style transfer, we also select two patches and enlarged them for better visualization. From the whole image of the stylized results (the first figure of each method), we can see the Neural-Style method fails to stylize the water. The whole image is a little dark, which may because the optimization-based method fail to find a good solution to balance the photorealistic content image and the textureless style image. The CNN-MRF method fails to stylize a smooth result, which may because that method requires a good correspondence between content image and style image. 
For the mountain peak (i.e., the second patch with blue border), we can see the result of our method is most similar to the one (i.e., blue bounding box) in the style image.
On the right column of Fig. \ref{fig:qua}, we compare our method with \cite{johnson_eccv16}. From the patches of tree and water, we can see our method presents styles more like ones in the style image.

Besides, we can see the stylized results generated by our method is sparser and have a high contrast, which are more similar to the style image.
These results verify the superiority of our method.

%The result of Fast-Neural-Style [] is comparable with ours, but our method is more ``abstract" and accommodate to the style of traditional Chinese painting. 
%Compare the leaves in the first patch (with red border) of each method, we can see our method presents less details and has a stronger constrast between the foreground and background, which is more like a Chinese painting.

\subsection{Ablation Study}

\begin{figure}
\centering
%\framebox
{\includegraphics[width = 1.0\textwidth]{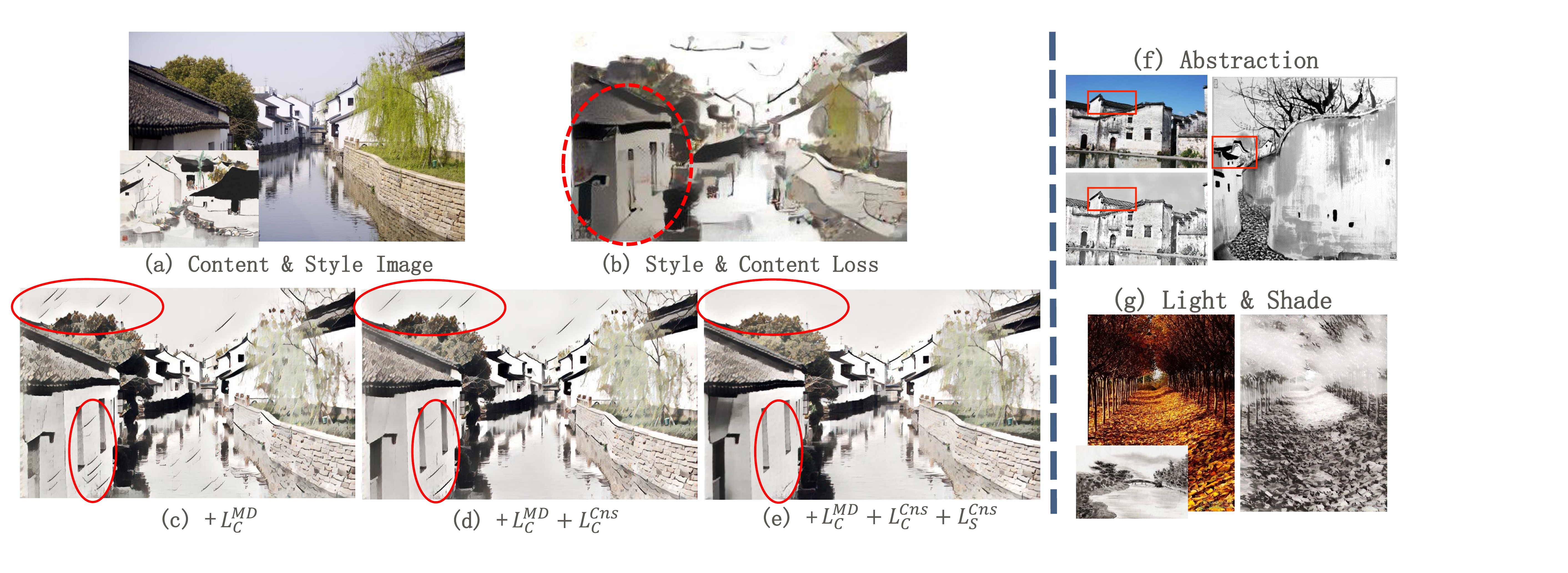}}
\caption{Left: Illustration of effects of our newly introduced losses. (a) style \& content images, (b) stylized result of \textit{our base model} that training with only content and style losses as Neural Style \cite{gatys_cvpr16}, (c,d,e) are the results generated by different variants of our model that trained with different MXDoG losses. Right: Two typical failure examples of our method. See text for details.}
%\caption{Illustration of effects of our newly introduced losses. (a) style \& content images, (b) our style transfer model with only content, style, and MXDoG content loss $L_C^{MD}$, (c) adding MXDoG content constraint loss $L_C^{Cns}$, (d) adding MXDoG content and style constraint losses $L_C^{Cns}$ and $L_S^{Cns}$.}
\label{fig:cns}
\end{figure} 

Fig. \ref{fig:cns} shows an illustration of effects of our three newly introduced losses, i.e., the  MXDoG content loss $L_C^{MD}$, the MXDoG content constraint loss $L_C^{Cns}$ and the MXDoG style constraint loss $L_S^{Cns}$ (\textit{we use a high style weight $\lambda_3 = 100.0$ for verification}). Fig. \ref{fig:cns}(a) shows the style and content images, Fig. \ref{fig:cns}(b) shows the result generated by \textit{our base model} that training with only style and content losses ($L_S$ and $L_C$), this similar to the Neural Style Transfer method \cite{gatys_cvpr16}.
The second column of Fig. \ref{fig:cns} shows the stylized results produced by different variants of our model. Specifically, Fig. \ref{fig:cns}(c) corresponds to the model that 
trained with only $L_C$, $L_S$, and  $L_C^{MD}$, Fig. \ref{fig:cns}(d) corresponds to the model that 
trained with  $L_C$, $L_S$,  $L_C^{MD}$, and $L_C^{Cns}$, Fig. \ref{fig:cns}(e) corresponds our model with full losses.
Compare with Fig. \ref{fig:cns}(b) and Fig. \ref{fig:cns}(c), we can see, with loss term $L_C^{MD}$, the stylized result is more natural and has a higher image contrast (e.g., the building in the dashed ellipse in Fig. \ref{fig:cns}(b) and Fig. \ref{fig:cns}(c)), but also introduce some artifacts. By adding loss functions of $L_C^{Cns}$ and $L_S^{Cns}$ sequentially, the result is more and more cleaned (refer to the solid ellipses in Fig. \ref{fig:cns}(c)(d)(e)).

\iffalse{
Fig. \ref{fig:cns} shows an illustration of effects of our three newly introduced losses, i.e., $L_C^{MD}$, $L_C^{Cns}$ and $L_S^{Cns}$ (we use a high style weight $10.0$ for verification). Fig. \ref{fig:cns}(a) shows the style and content image, Fig. \ref{fig:cns}(b) shows the result of our model that training with only style, content and MXDoG content losses $L_C^{MD}$. Fig. \ref{fig:cns}(c) shows the result of our method by adding MXDoG content constraint loss $L_C^{Cns}$, and Fig. \ref{fig:cns}(e) shows the result of our method with full losses. From which we can see, with loss function $L_C^{MD}$, we can make the stylized result abstract but may also introduce some artifacts (Fig. \ref{fig:cns}(b)). By adding loss functions of $L_C^{Cns}$ and $L_S^{Cns}$, the result is more and more cleaned (Fig. \ref{fig:cns}(c)(d)).
}\fi

%$L_C^{Cns}$ and $L_S^{Cns}$ are mainly used to clean the large noisy edges in the stylized image. As we care more about the abstract parts insteading of the very low-level details in an image, our modified XDoG is more preferable than the Laplacian filter \cite{lapstyle} in Chinese traditional painting transfer.

\begin{figure*}
\centering
%\framebox
{\includegraphics[width = 1.0\textwidth]{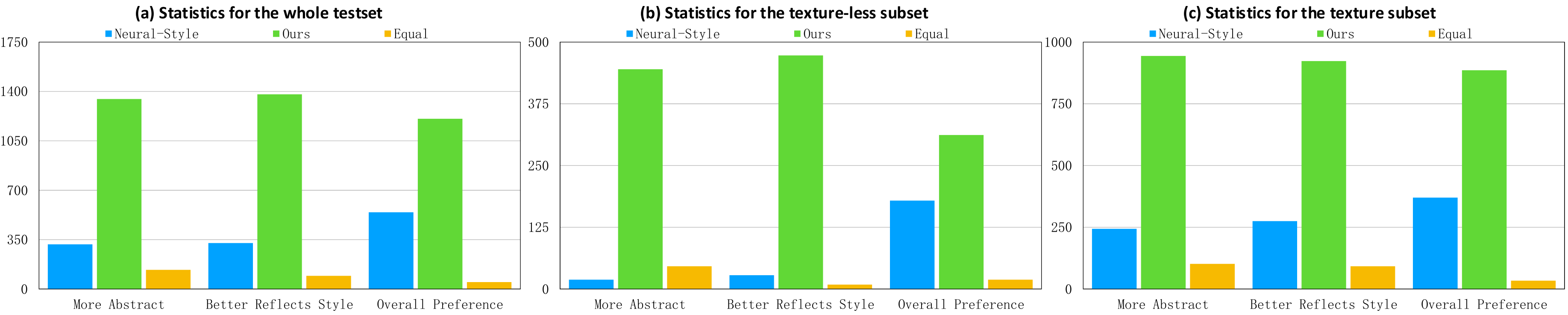}}
\caption{Results of user study. For each group, the blue bar shows the votes received by Fast-Neural-Style, the green bar shows the votes of our method, and the yellow bar shows the votes of ``Equally good or undecided". }
\label{fig:user}
\end{figure*} 

\subsection{User Study}
We carry out a user study to quantitatively evaluate the proposed method. We first randomly select $10$ styles of typical Chinese traditional painting, then randomly select $10$ stylized images for each style. We invite $60$ people, aged from 21 to 45, with diverse educational backgrounds, to participate in our study, each person is asked to cope with $30$ randomly selected stylized images, resulting a total of $1,800$ trials.
In this user study, we compare with the Fast-Neural-Style  \cite{johnson_eccv16} which can be seen as the fast version of \cite{gatys_cvpr16}. In each trial, a user is showed with the original image, the style image, and results of  \cite{johnson_eccv16} and our model.  The order of the stylized results is randomized to avoid participants' laziness.
For each person, we ask three questions: ``Which of the two stylized results is more abstract?", ``Which of the two stylized results better reflects the style of the painting?", and ``Overall, which of the two stylized results do you prefer?". Participants have to select either one of the stylized results or ``Equally good or undecided".
Overall results of the user study are showed in Fig. \ref{fig:user}(a), which indicating a clear preference of our method.
%This user study verifies the effectiveness of our method on transferring the style of Chinese traditional painting.

However, as can be seen in Fig. \ref{fig:user}(a), there are still some people prefer the result of Fast-Neural-Style \cite{johnson_eccv16}, thus we make a detailed study about the style patterns. 

We first analyse patterns for which the result of our method is preferred by users. We find our model works better on abstract, textureless and less colorful styles. Specifically, it can drop out some tedious details and capture the essence of scenes or objects, which making the whole stylized images looks concise and clean. 
To verify our observation, we further split the stylized results by the extent of texture and color of style images\footnote{For instance, figures in the first row of Fig. \ref{fig:dataset} represent typical examples of textured and colorful style images, while figures in the second row of Fig. \ref{fig:dataset} stand for the texture-less and less colorful styles.}. Fig \ref{fig:user}(b) shows the statistics on results of less textured and colorful styles, and Fig \ref{fig:user}(c) shows the statistics on results of textured and colorful styles. 

For votes on ``More Abstract" and ``Better Reflects Style", we can see stylized results of our method are more preferred on transferring less textured and colorful styles. We think this is because more details in the photo are expected to be discarded to match the style of the style image.

However, the preponderance of our method on ``Overall Preference" is shrunk, which indicates that although some people think our stylized results are more abstract and closer to the style image, they still like the more concrete images generated by the Fast-Neural-Style. This reflects some inconsistence of ``abstraction" and ``aesthetical-appealing", and also more powerful method is needed for abstraction, as we can see, the leaves and tree branches on the right of figure \ref{fig:qua} is still not good enough as the style image. 

%Second, we find that there are some Chinese traditional painting that have relatively rich texture and color information, e.g., the fourth figure in the first row of \ref{fig:dataset}. In this case, more details in the photo need to be reserved, and differences of our method and Fast-Neural-Style are scaled down.

\subsection{Failure Examples}
Although our method works well with the general  Chinese traditional painting style transfer, we find it still could not achieve the result of a trained human artist in case of ``abstraction" and handling the ``light and shade".
%Fig. \ref{fig:rst_future} shows two examples. 
Fig. \ref{fig:cns}(f) shows our model fails to  find the correct correspondence of roof between the content image and the style image.
%, i.e.,  the roof in the stylized result seems to match with the ground instead of the roof in the style image. This mismatching is due to the loss network is pretrained on the object recognition task, which is not aware of the roof.
Besides, the roofs in the stylized result should have black colors and curving shapes as the one in the style image, instead of rigid shapes chequered with black and white colors. 
We think the deep reason is that the neural network is still lack of the human-level ``abstraction" capability. Some research \cite{xdog} show ``abstraction" may have strong relevance with the semantic correspondence, thus we believe training a good loss network that can recognize objects and scenes in both photorealistic and artistic images will be a good promising direction.
Fig. \ref{fig:cns}(g) shows our method also fails to stylize images with alternating light and shade. Experiments indicate this is a common failure for all the neural style transfer methods and we leave it as an interesting future work.

\iffalse{
\begin{figure}
\centering
%\framebox
{\includegraphics[width = 0.86\textwidth]{./rst_future4.pdf}}
\caption{Two typical failure examples of our method.}
\label{fig:rst_future}
\end{figure} 
}\fi

\section{Discussion and Conclusion} 
Chinese traditional painting is very popular in Eastern Asia, the style of which is often abstract and textureless. This is very different from Western Oil painting, and is not well transferred by current neural style transfer methods \cite{gatys_cvpr16,johnson_eccv16,lichuan1}.
To tackle this problem, we propose a novel neural style transfer method for Chinese traditional painting. We first introduce a MXDoG filter, then incorporate the MXDoG function with three new loss terms for network training. The effects of our method are verified on test images with diverse conditions.
To further promote this research direction, we introduce the Chinese traditional painting dataset which containing diverse content and style images to the public.

Although our method shows superiorities on ``abstraction" and in accordance with Chinese traditional painting over current neural style methods, it should be pointed out that the  abstraction and aesthetics produced by our model has  limitations and does not compete to the one produced by a trained artist. 
For example, The tree branches stylized by our model in Fig. \ref{fig:qua} is still not comparable with  ones in the style image. However, as {\it abstraction} remains some of the fundamentally unsolved problems in non-photorealistic rendering (NPR) \cite{xdog}, this arguably good results still  help a lot for our neural style model to get a freehand painting and might steer deeper research into the artistic neural style transfer community.

\noindent\textbf{Acknowledgement:} T. Wu is supported by ARO award W911NF1810295, ARO DURIP award W911NF1810209 and NSF IIS 1822477. 

\newpage
%===========================================================
\bibliographystyle{splncs04}
\bibliography{transfer}

\begin{thebibliography}{10}
\providecommand{\url}[1]{\texttt{#1}}
\providecommand{\urlprefix}{URL }
\providecommand{\doi}[1]{https://doi.org/#1}

\bibitem{artisto}
{Artisto}: http://artisto.my.com/ (2016)

\bibitem{baxter2001}
Baxter, B.: Dab: interactive haptic painting with 3d virtual brushes. In: ACM
  SIGGRAPH 2001 video review on Animation theater program. p.~10 (2001)

\bibitem{museum}
Becattini, F., Ferracani, A., Landucci, L., Pezzatini, D., Uricchio, T.,
  Del~Bimbo, A.: Imaging Novecento. A Mobile App for Automatic Recognition of
  Artworks and Transfer of Artistic Styles (2016)

\bibitem{chen2017coherent}
Chen, D., Liao, J., Yuan, L., Yu, N., Hua, G.: Coherent online video style
  transfer. In: Proc. Intl. Conf. Computer Vision (ICCV) (2017)

\bibitem{chen2017stylebank}
Chen, D., Yuan, L., Liao, J., Yu, N., Hua, G.: Stylebank: An explicit
  representation for neural image style transfer. In: Proc. CVPR (2017)

\bibitem{chen2018stereoscopic}
Chen, D., Yuan, L., Liao, J., Yu, N., Hua, G.: Stereoscopic neural style
  transfer. Proc. CVPR  (2018)

\bibitem{gatys_cvpr16}
Gatys, L.A., Ecker, A.S., Bethge, M.: Image style transfer using convolutional
  neural networks. In: Proceedings of the IEEE Conference on Computer Vision
  and Pattern Recognition (2016)

\bibitem{gatys_cvpr17}
Gatys, L.A., Ecker, A.S., Bethge, M., Hertzmann, A., Shechtman, E.: Controlling
  perceptual factors in neural style transfer. In: Proceedings of the IEEE
  Conference on Computer Vision and Pattern Recognition (Jul 2017)

\bibitem{style_wenhan}
Huang, H., Wang, H., Luo, W., Ma, L., Jiang, W., Zhu, X., Li, Z., Liu, W.:
  Real-time neural style transfer for videos. In: CVPR (2017)

\bibitem{johnson_eccv16}
Johnson, J., Alahi, A., Fei-Fei, L.: Perceptual losses for real-time style
  transfer and super-resolution. In: European Conference on Computer Vision
  (2016)

\bibitem{kristen}
Joshi, B.J., Stewart, K., Shapiro, D.: Bringing impressionism to life with
  neural style transfer in come swim. CoRR  \textbf{abs/1701.04928} (2017)

\bibitem{adam}
Kingma, D.P., Ba, J.: Adam: A method for stochastic optimization. CoRR
  \textbf{abs/1412.6980} (2014)

\bibitem{lecun98}
Lecun, Y., Bottou, L., Bengio, Y., Haffner, P.: Gradient-based learning applied
  to document recognition. In: Proceedings of the IEEE. pp. 2278--2324 (1998)

\bibitem{lee1999simulating}
Lee, J.: Simulating oriental black-ink painting. Computer Graphics \&
  Applications IEEE  \textbf{19}(3),  74--81 (1999)

\bibitem{lichuan1}
Li, C., Wand, M.: Combining markov random fields and convolutional neural
  networks for image synthesis. CoRR  \textbf{abs/1601.04589} (2016)

\bibitem{lichuan2}
Li, C., Wand, M.: Precomputed real-time texture synthesis with markovian
  generative adversarial networks. CoRR  \textbf{abs/1604.04382} (2016)

\bibitem{lapstyle}
Li, S., Xu, X., Nie, L., Chua, T.S.: Laplacian-steered neural style transfer.
  In: Proceedings of the ACM Multimedia Conference (MM). (2017)

\bibitem{Chinese_painting_lindaoyu}
Lin, D., Wang, Y., Xu, G., Li, J., Fu, K.: Transform a simple sketch to a
  chinese painting by a multiscale deep neural network. Algorithms
  \textbf{11}(1), ~4 (2018)

\bibitem{coco}
Lin, T., Maire, M., Belongie, S.J., Bourdev, L.D., Girshick, R.B., Hays, J.,
  Perona, P., Ramanan, D., Doll{\'{a}}r, P., Zitnick, C.L.: Microsoft {COCO:}
  common objects in context. CoRR  \textbf{abs/1405.0312} (2014)

\bibitem{auto_painter}
Liu, Y., Qin, Z., Luo, Z., Wang, H.: Auto-painter: Cartoon image generation
  from sketch by using conditional generative adversarial networks. CoRR
  \textbf{abs/1705.01908} (2017)

\bibitem{luan2017deep}
Luan, F., Paris, S., Shechtman, E., Bala, K.: Deep photo style transfer. arXiv
  preprint arXiv:1703.07511  (2017)

\bibitem{edge_theory}
Marr, D., Hildreth, E.: Theory of edge detection. {P}roceedings of the Royal
  Society of {L}ondon Series {B}  \textbf{207},  187--217 (1980)

\bibitem{nikulin1}
Nikulin, Y., Novak, R.: Exploring the neural algorithm of artistic style. CoRR
  \textbf{abs/1602.07188} (2016)

\bibitem{nikulin2}
Novak, R., Nikulin, Y.: Improving the neural algorithm of artistic style. CoRR
  \textbf{abs/1605.04603} (2016)

\bibitem{pytorch}
Paszke, A., Gross, S., Chintala, S., Chanan, G., Yang, E., DeVito, Z., Lin, Z.,
  Desmaison, A., Antiga, L., Lerer, A.: Automatic differentiation in pytorch
  (2017)

\bibitem{prisma}
{Prisma}: http://prisma-ai.com/ (2016)

\bibitem{imagenet}
Russakovsky, O., Deng, J., Su, H., Krause, J., Satheesh, S., Ma, S., Huang, Z.,
  Karpathy, A., Khosla, A., Bernstein, M., Berg, A.C., Fei-Fei, L.: {ImageNet
  Large Scale Visual Recognition Challenge}. International Journal of Computer
  Vision (IJCV)  \textbf{115}(3),  211--252 (2015)

\bibitem{portrait}
Selim, A., Elgharib, M., Doyle, L.: Painting style transfer for head portraits
  using convolutional neural networks. ACM Trans. Graph.  \textbf{35}(4),
  129:1--129:18 (Jul 2016)

\bibitem{vgg}
Simonyan, K., Zisserman, A.: Very deep convolutional networks for large-scale
  image recognition. CoRR  \textbf{abs/1409.1556} (2014)

\bibitem{strassmann}
Strassmann, S.: Hairy brushes. In: Conference on Computer Graphics and
  Interactive Techniques. pp. 225--232 (1986)

\bibitem{ulyanov}
Ulyanov, D., Lebedev, V., Vedaldi, A., Lempitsky, V.S.: Texture networks:
  Feed-forward synthesis of textures and stylized images. CoRR
  \textbf{abs/1603.03417} (2016)

\bibitem{inst_norm}
Ulyanov, D., Vedaldi, A., Lempitsky, V.S.: Instance normalization: The missing
  ingredient for fast stylization. CoRR  \textbf{abs/1607.08022} (2016)

\bibitem{way2002}
Way, D.L., Lin, Y.R., Shih, Z.C.: The synthesis of trees in chinese landscape
  painting using silhouette and texture strokes. Journal of Wscg  \textbf{10},
  499--506 (2002)

\bibitem{xdog}
Winnemöller, H., Kyprianidis, J.E., Olsen, S.C.: Xdog: An extended
  difference-of-gaussians compendium including advanced image stylization.
  Computers \& Graphics  \textbf{36}(6),  740 -- 753 (2012)

\bibitem{xu2006animating}
Xu, S., Xu, Y., Kang, S.B., Salesin, D.H., Pan, Y., Shum, H.Y.: Animating
  chinese paintings through stroke-based decomposition. ACM Transactions on
  Graphics  \textbf{25}(2),  239--267 (2006)

\bibitem{zhangsonghai}
Zhang, S.H., Chen, T., Zhang, Y.F., Hu, S.M., Martin, R.: Video-based running
  water animation in chinese painting style. Science China Information Sciences
   \textbf{52}(2),  162--171 (2009)

\bibitem{filamentary}
Zhao, H., Li, H., Cheng, L.: Synthesizing filamentary structured images with
  gans. CoRR  \textbf{abs/1706.02185} (2017)

\bibitem{cycle_gan}
Zhu, J.Y., Park, T., Isola, P., Efros, A.A.: Unpaired image-to-image
  translation using cycle-consistent adversarial networks. arXiv preprint
  arXiv:1703.10593  (2017)

\end{thebibliography}

\end{document}